\pdfoutput=1
\documentclass[11pt]{article}

\usepackage[preprint]{acl}

\usepackage{times}
\usepackage{latexsym}

\usepackage[T1]{fontenc}
\usepackage[utf8]{inputenc}

\usepackage{microtype}

\usepackage{inconsolata}

\usepackage{graphicx}
\usepackage{amsmath}
\usepackage{subfig}
\usepackage{float}
\usepackage{multirow}
\usepackage{makecell}
\usepackage{booktabs}
\usepackage{amsfonts}
\usepackage{colortbl}

\title{BeamLoRA: Beam-Constraint Low-Rank Adaptation}
\author{Naibin Gu\textsuperscript{\rm 1,2}\thanks{\ \  Equal contribution.}\thanks{\ \  This work was done during an internship at Baidu Inc.},\ Zhenyu Zhang\textsuperscript{\rm 3}\footnotemark[1],\ Xiyu Liu\textsuperscript{\rm 1,2},\ Peng Fu\textsuperscript{\rm 1,2}\thanks{\ \  Corresponding author: Peng Fu.},\ Zheng Lin\textsuperscript{\rm 1,2},\\ {\bf Shuohuan Wang\textsuperscript{\rm 3},\ Yu Sun\textsuperscript{\rm 3},\ Hua Wu\textsuperscript{\rm 3},\ Weiping Wang\textsuperscript{\rm 1},\ Haifeng Wang\textsuperscript{\rm 3}} \\ 
\textsuperscript{\rm 1}Institute of Information Engineering, Chinese Academy of Sciences, Beijing, China \\
\textsuperscript{\rm 2}School of Cyber Security, University of Chinese Academy of Sciences, Beijing, China \\
\textsuperscript{\rm 3}Baidu Inc., Beijing, China \\
  \texttt{ \textrm{\{}gunaibin,fupeng\textrm{\}}@iie.ac.cn} \\
  \texttt{ \textrm{\{}zhangzhenyu07,wangshuohuan\textrm{\}}@baidu.com} \\
}

\begin{document}
\maketitle
\begin{abstract}
Due to the demand for efficient fine-tuning of large language models, Low-Rank Adaptation (LoRA) has been widely adopted as one of the most effective parameter-efficient fine-tuning methods. Nevertheless, while LoRA improves efficiency, there remains room for improvement in accuracy. Herein, we adopt a novel perspective to assess the characteristics of LoRA ranks. The results reveal that different ranks within the LoRA modules not only exhibit varying levels of importance but also evolve dynamically throughout the fine-tuning process, which may limit the performance of LoRA. Based on these findings, we propose BeamLoRA, which conceptualizes each LoRA module as a beam where each rank naturally corresponds to a potential sub-solution, and the fine-tuning process becomes a search for the optimal sub-solution combination. BeamLoRA dynamically eliminates underperforming sub-solutions while expanding the parameter space for promising ones, enhancing performance with a fixed rank. Extensive experiments across three base models and 12 datasets spanning math reasoning, code generation, and commonsense reasoning demonstrate that BeamLoRA consistently enhances the performance of LoRA, surpassing the other baseline methods\footnote{The code is available at \url{https://github.com/gccnlp/BeamLoRA}.}.
\end{abstract}

\section{Introduction}
In recent years, large language models have shown tremendous potential in various applications~\citep{Touvron2023LLaMAOA, DBLP:journals/corr/abs-2307-09288, DBLP:journals/corr/abs-2310-06825, OpenAI2023GPT4TR,chen2024mixturehiddendimensionstransformer,qwen2025qwen25technicalreport}. To further enhance model performance on specific downstream tasks, fine-tuning is usually the most effective approach. However, as the scale of models keeps increasing, fine-tuning all model parameters becomes unsustainable. To address this issue, parameter-efficient fine-tuning (PEFT) emerges as a practical solution~\citep{DBLP:conf/icml/HoulsbyGJMLGAG19,li-liang-2021-prefix,liu-etal-2022-p,DBLP:conf/iclr/HuSWALWWC22, gu-etal-2024-light, jiang2024morahighrankupdatingparameterefficient}. By updating only lightweight modules, these methods nearly achieve the results of full parameter fine-tuning while reducing both fine-tuning time and memory usage. 

Among these PEFT methods, Low-Rank Adaptation~(LoRA) stands out for its effectiveness and practicality~\citep{DBLP:conf/iclr/HuSWALWWC22}. The method strategically inserts trainable low-rank modules into frozen linear layers, approximating weight updates while preserving the original model architecture and inference efficiency. 
Recent advancements aim to enhance LoRA through various approaches: DoRA~\citep{pmlr-v235-liu24bn} decouples the fine-tuning process into directional and magnitude adjustments, whereas AdaLoRA~\citep{zhang2023adaptive} and IncreLoRA~\citep{DBLP:journals/corr/abs-2308-12043} dynamically optimize rank allocation across different modules.  However, when revisiting the fundamental aspects of LoRA, we find these methods generally treat rank dimensions as homogeneous units, neglecting the potential hierarchical importance of individual rank components within each LoRA module. 

In this paper, we adopt a novel perspective by studying the intrinsic characteristics of LoRA ranks from both spatial and temporal dimensions. From the spatial dimension, we find significant differences in the importance of ranks within a LoRA module, and pruning the less important ranks has a minimal impact on performance. From the temporal dimension, these important differences do not show up at the beginning of fine-tuning, but gradually emerge and stabilize as the fine-tuning process progresses. Despite the significant differences in importance among ranks, existing works typically allocate the same parameter budget to each rank (i.e., a row and a column in a module), which leads to constrained optimization space for important ranks and wasted resources on less important ones. 

Based on the spatial and temporal findings, we propose BeamLoRA, which is inspired by beam search~\citep{lowerre1976harpy} and treats each LoRA module as a beam, where each rank acts as a sub-solution, and the fine-tuning process is formalized as searching for the optimal combination of sub-solutions. 
Specifically, the main process of BeamLoRA includes assessment, pruning, and expansion. To assess the importance of each sub-solution, we insert a trainable score vector into the low-rank subspace and integrate the assessment process into fine-tuning. Based on their importance, we prune unimportant sub-solutions to free up space and expand the important ones, thereby allowing them to be better optimized. Furthermore, to better determine the pruning or expansion threshold, we introduce a dynamic Top-P method that achieves adaptability in both temporal and spatial dimensions. Through these mechanisms, BeamLoRA can effectively allocate parameter capacity to the most promising solution paths.

We validate our approach using three different base models across 12 datasets covering math reasoning, code generation, and commonsense reasoning. Results indicate that BeamLoRA consistently outperforms multiple LoRA-enhanced baselines. Notably, on the most challenging math reasoning and code generation tasks, BeamLoRA achieves a 1.57\% accuracy gain while using only 2.4\% of the trainable parameters compared to full fine-tuning. 
Further analysis reveals that the success of BeamLoRA is attributed to its increased important rank space within the LoRA module.

In summary, our contributions are as follows:
\begin{itemize}
\item We adopt a novel perspective by studying the characteristics of LoRA ranks from both spatial and temporal dimensions, and highlight that ranks with various importance are assigned an equally sized parameter space.
\item We introduce BeamLoRA and view a LoRA module as a beam. It continuously assesses the importance of each rank, compresses the less important ones, and frees up resources for the more significant ones.
%  with each rank as a sub-solution During fine-tuning, 
\item Through extensive experiments across three base models of different sources and scales, along with 12 diverse datasets, we demonstrate that BeamLoRA consistently outperforms other baselines.
\end{itemize}
\section{Preliminary}
\subsection{Low-Rank Adaptation (LoRA)}
Considering that the updates for fine-tuning large models occur within a low-rank subspace~\citep{aghajanyan-etal-2021-intrinsic}, LoRA inserts low-rank modules into the linear layers of the base model to approximate these transformations. Specifically, for a weight matrix $\mathbf{W}_{0}\in{\mathbb{R}^{d\times k}}$, LoRA decomposes the update $\Delta \mathbf{W}$ into a low-rank matrices product $\mathbf{BA}$, where $\mathbf{B}\in{\mathbb{R}^{d\times r}}$, $\mathbf{A}\in{\mathbb{R}^{r\times k}}$, and $r\ll min(d,k)$. The forward pass of LoRA is formulated as
\begin{equation}
    y= \mathbf{W}_{0}x+\Delta \mathbf{W}x=\mathbf{W}_{0}x+\mathbf{BA}x,
\end{equation}
where $x$ represents the input and $y$ is the output. During fine-tuning, $\mathbf{W}_{0}$ remains frozen, while only $\mathbf{B}$ and $\mathbf{A}$ matrices are trainable.

\noindent\textbf{The Independence of Ranks.} Given a LoRA module that includes two matrices $\mathbf{B}$ and $\mathbf{A}$, in which $\mathbf{B}$ is represented as $[\mathbf{b}_{1}, \mathbf{b}_{2},...,\mathbf{b}_{r}]$, where $\mathbf{b}_{i}$ denotes the $i$-th column of matrix $\mathbf{B}$, and $\mathbf{A}=[\mathbf{a}_{1},\mathbf{a}_{2},...,\mathbf{a}_{r}]$, where $\mathbf{a}_{i}$ denotes the $i$-th row. In this way, the update $\Delta \mathbf{W}$ is equivalent to
\begin{equation}
\begin{aligned}
    \Delta \mathbf{W}=& \mathbf{BA}= \begin{bmatrix}\mathbf{b}_1 \; \mathbf{b}_2 \; ... \; \mathbf{b}_r\end{bmatrix} \begin{bmatrix}\mathbf{a}_1 \\ \mathbf{a}_2 \\ ... \\ \mathbf{a}_r\end{bmatrix}, \\
    = & \mathbf{b}_1\mathbf{a}_1+...+\mathbf{b}_r\mathbf{a}_r =\sum_r{\mathbf{b}_i\mathbf{a}_i} =\sum_r{\Delta \mathbf{w}_i},
\end{aligned}
\end{equation}
where $\Delta \mathbf{w}_i\in{\mathbb{R}^{d\times k}}$ represents the update matrix of $i$-th rank. Thus, the LoRA fine-tuning process can be viewed as independently updating each $\Delta \mathbf{w}_i$ represented by each rank.

\subsection{Analysis of LoRA Ranks}
\label{pilot}
During the fine-tuning process, an intuitive assumption is that each rank within a LoRA module contributes similarly. This intuition may stem from the standard LoRA initialization procedure, where matrix $\mathbf{A}$ is initialized randomly, and matrix $\mathbf{B}$ starts with zero values. Since all $\Delta \mathbf{w}_i$ matrices begin as zero matrices and are updated simultaneously, their contributions might remain comparable throughout the fine-tuning process. 

To examine the validity of this assumption, we fine-tune LoRA on LLaMA2-7B~\citep{DBLP:journals/corr/abs-2307-09288} and Mistral-7B-v0.1~\citep{DBLP:journals/corr/abs-2310-06825} with the MetaMathQA dataset~\citep{yu2024metamath} and conduct an analysis from both spatial and temporal dimensions. Given that LoRA updates represent adjustments to pre-trained weights, we use the magnitude of each $\Delta \mathbf{w}$ to quantify the importance of different ranks\footnote{The magnitude (importance) of the matrix is roughly measured by the commonly used Frobenius norm.}.

\begin{figure}[t]
        \centering
	\subfloat[LLaMA2-7B]{
	\includegraphics[width=0.48\linewidth]{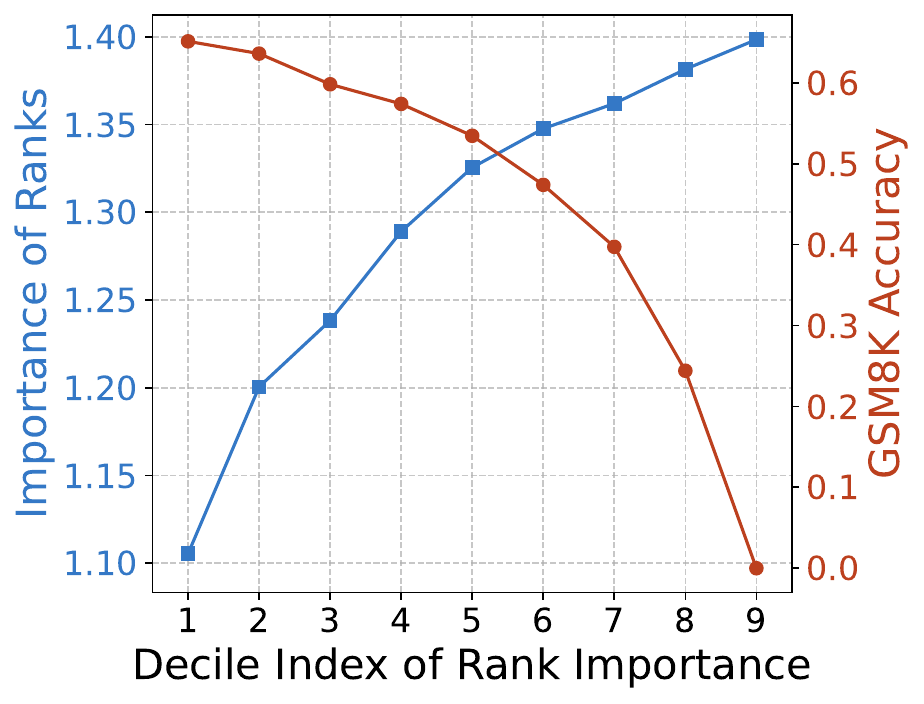}} 
        \subfloat[Mistral-7B-v0.1]{
	\includegraphics[width=0.48\linewidth]{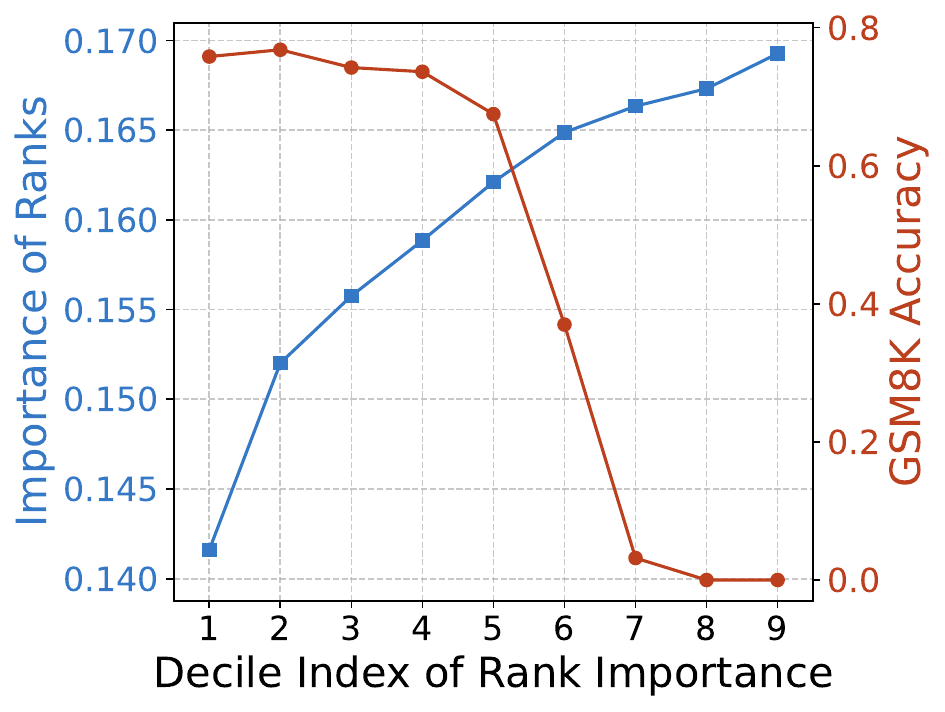}} 
\caption{Differences in importance among ranks within a LoRA module (\emph{spatial}). The blue line represents the deciles of importance for ranks. The red line represents accuracy changes when pruning ranks of varying importance. We take \texttt{ffn.up\_proj} in the 30th layer as an example, with similar phenomena in other modules. By pruning ranks, it can be observed that pruning less important ranks has a minor effect, while pruning more important ranks has a significant impact.}
\label{PilotA}
\end{figure}

\noindent\textbf{Spatial Dimension.}  
Figure~\ref{PilotA} describes the results of sorting the importance of ranks after fine-tuning. The blue line represents the levels of importance across different deciles after being sorted by importance in a LoRA module. It can be observed that the deciles of the importance of ranks are hierarchical, indicating that \textbf{the importance of different ranks within a LoRA module is not uniform after fine-tuning.} Furthermore, by pruning the ranks in each LoRA module based on its own importance from the least to the most significant gradually (the red line), \textbf{the accuracy shows limited change when pruning the less important ranks}. On the contrary, when important ranks are pruned, the evaluation results drop sharply to zero. This phenomenon further demonstrates the significant differences in importance among the different ranks.
\begin{figure}[t]
        \centering
	\subfloat[\texttt{ffn.up\_proj}]{
	\includegraphics[width=0.46\linewidth]{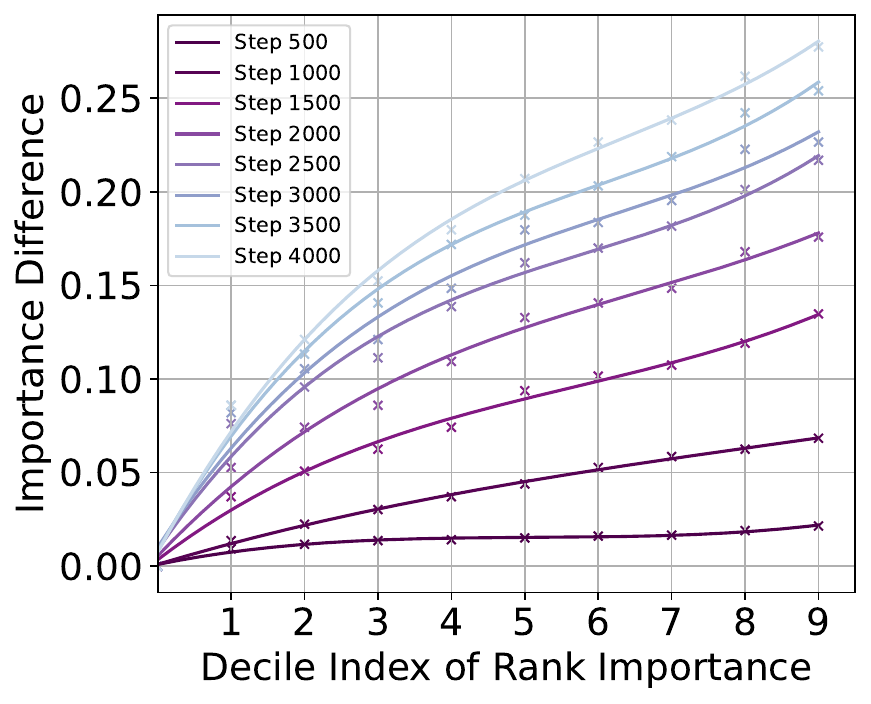}} 
        \subfloat[\texttt{attn.v\_proj}]{
	\includegraphics[width=0.46\linewidth]{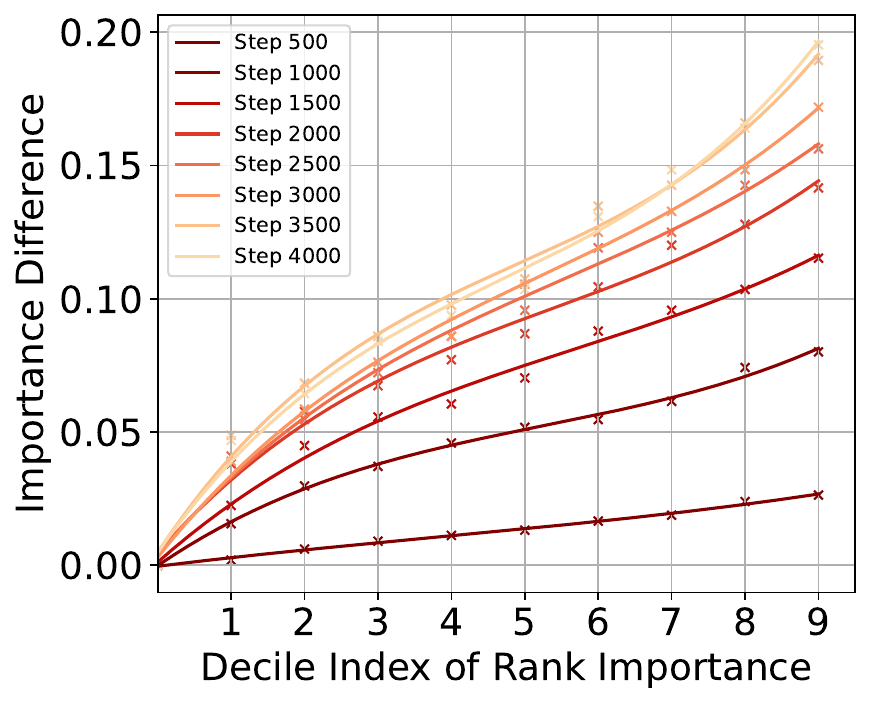}} 
\caption{The visualization of importance difference among different ranks in LoRA on LLaMA2-7B as fine-tuning steps increase (\emph{temporal}). We take \texttt{ffn.up\_proj} and \texttt{attn.v\_proj} in the 30th layer as examples, with similar trends observed in all other modules. It can be observed that the importance of ranks is equal at the beginning of fine-tuning, but differences gradually emerge as the fine-tuning process progresses.}
\label{PilotB}
\end{figure}

\noindent\textbf{Temporal Dimension.} 
To further understand the reasons behind this importance differentiation, we return to the initial assumption: since the $\Delta \mathbf{w}$ corresponding to each rank is initialized to zero, they all start with equal importance during fine-tuning. Therefore, a natural idea is to investigate how the importance of different ranks evolves during fine-tuning. Figure~\ref{PilotB} shows the changes in importance of two LoRA modules, where \textbf{the differences in importance among ranks increase with the number of fine-tuning steps}. In other words, the less important ranks are progressively filtered out. Furthermore, \textbf{the differences in importance tend to stabilize as the number of fine-tuning steps continues to increase further}. These phenomena are prevalent across various LoRA modules. 

In summary, there are significant differences in the importance of ranks in LoRA, and these differences appear and gradually increase as the fine-tuning process progresses. However, in most existing LoRA-based methods, less important ranks still occupy the same parameter budget as important ones. Here, a question is about to arise: 

\textit{Could we free up space from less important ranks for more important ones to achieve better optimization?}

\begin{figure*}[t]
    \centering
    \includegraphics[width=2\columnwidth,trim=52 50 52 50,clip]{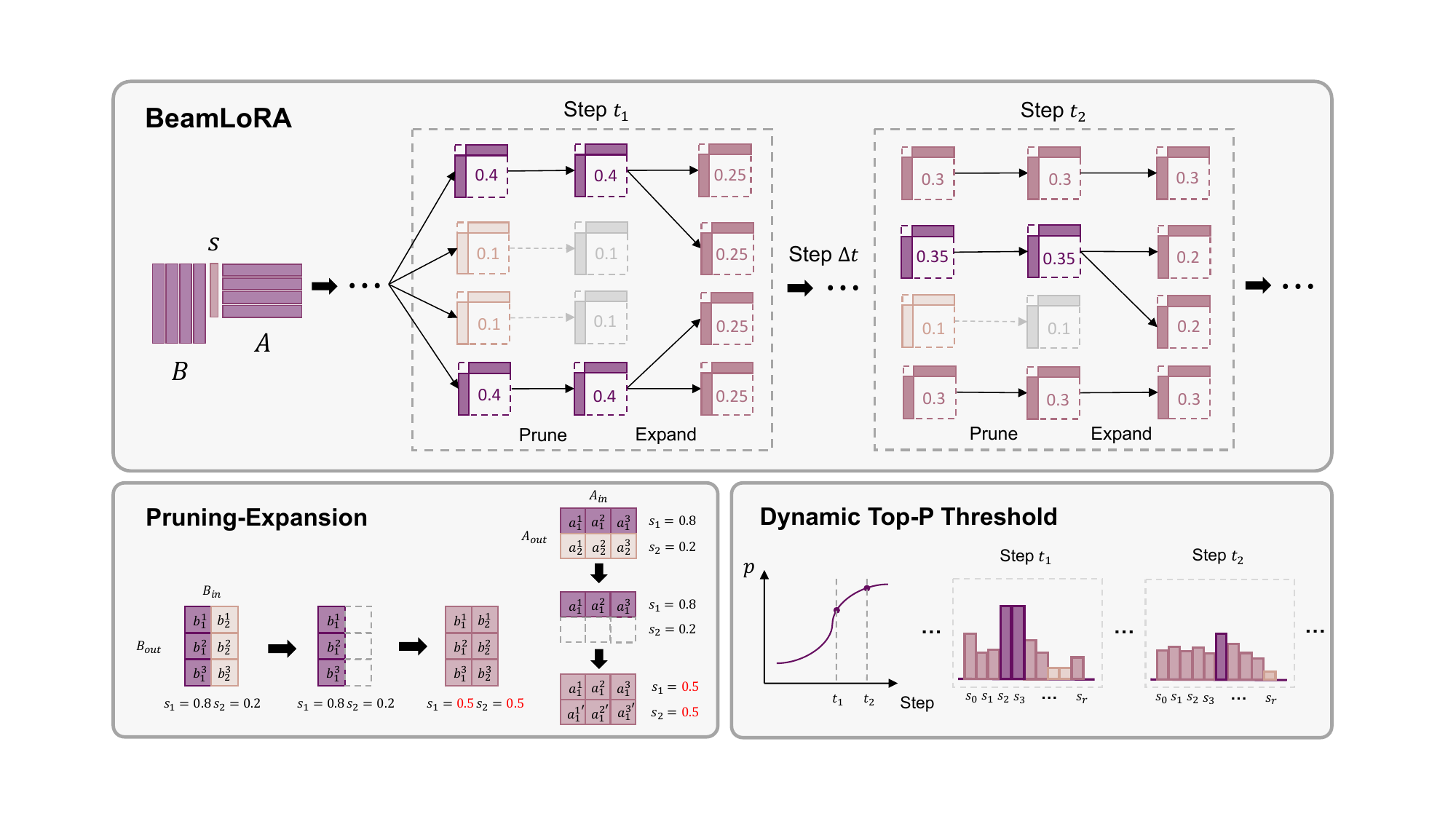}
    \caption{Illustration of BeamLoRA. Throughout the fine-tuning process, BeamLoRA continually assesses the importance of each rank. Every $\Delta t$ steps, unimportant ranks are pruned while those identified as important are expanded, optimizing the module’s performance.}
    \label{method}
\end{figure*}
\section{BeamLoRA}
To answer the above question, we propose BeamLoRA, which continuously assesses the importance of different ranks during fine-tuning, periodically pruning the less important ones to free up resources for the more important ranks. The overall workflow of the method is illustrated in Figure~\ref{method}.

For a LoRA module with rank $r$,
% given a fixed parameter budget $r$, 
we treat it as a beam with width $r$ and the optimization process is naturally regarded as a search for the solution set $\Delta \mathbf{W}=\{\Delta \mathbf{w}_1,\Delta \mathbf{w}_2,...,\Delta \mathbf{w}_r\}$ tailored to the fine-tuning dataset, where the $i$-th rank in the LoRA module is considered a sub-solution  $\Delta \mathbf{w}_i$. Formally, the optimization process seeks to minimize the loss function $\mathcal{L}$ over the dataset $\mathcal{D}$:
\begin{equation}
    \Delta \mathbf{W}^* = \arg\min\ \mathcal{L}(\mathbf{W}_0+\Delta \mathbf{W};\ \mathcal{D}),
\end{equation}
where the optimal solution $\Delta \mathbf{W}^*$ represents the well-trained LoRA module. 

\subsection{Importance Assessment}
In the pilot experiments of Section~\ref{pilot}, we use the Frobenius norm to measure the importance of each rank offline. However, this approach involves considerable computational overhead during fine-tuning\footnote{Typically, a large model contains hundreds of LoRA modules. For each module, it requires computing $r$ matrices of size $d\times k$, and then calculating the norm for each matrix.}. 
To make the assessment more efficient and accurate, we introduce a learnable score vector $\mathbf{s}\in{\mathbb{R}^r}$, which is inserted between the matrices $\mathbf{B}$ and $\mathbf{A}$, to scale the output of each rank through element-wise broadcasting multiplication. In that case, the modified forward pass of LoRA can be formulated as follows:
\begin{equation}
\begin{aligned}
    y=\mathbf{W}_{0}x+\mathbf{B}(\mathbf{s}\odot \mathbf{A})x=\sum_r&{s_i\Delta \mathbf{w}_ix}.
\end{aligned}
\end{equation}
This is equivalent to scaling the corresponding rank matrices $\Delta \mathbf{w}_i$. 
This means that during fine-tuning, if a rank is considered important, the corresponding score $s_i$ for that rank is amplified.

At the start of the fine-tuning process, LoRA initializes each rank to zero, indicating that their initial importance is equal. Consequently, we initialize all elements in $\mathbf{s}$ with identical values. During the fine-tuning process, $\mathbf{s}$ is consistently normalized using the softmax function, like the logits of tokens in text generation:
\begin{equation}
\begin{aligned}
s_i =  \frac{e^{s_i}}{\sum_{j=1}^{r} e^{s_j}},
\end{aligned}
\end{equation}
where $s_i$ is the $i$-th element in score vector $\mathbf{s}$. The continuous normalization ensures a stable value range and facilitates meaningful comparisons of importance differences between elements.

\subsection{Pruning and Expansion}
With the importance of each rank, the space occupied by the less important ranks can be freed up, which allows us to expand the parameter space for the remaining important ones. Specifically, we begin by selecting the $K$ least important ranks based on their importance $\mathbf{s}$ to form the rank index set $\mathcal{I}_p$ for pruning:
\begin{equation}
\begin{aligned}
\label{eq:Iu}
\mathcal{I}_p=\{i\ | s_i \in \text{Min}_K(\mathbf{s})\}.
\end{aligned}
\end{equation}
During the pruning stage, for the indices of the unimportant ranks included in $\mathcal{I}_p$, we set their parameters to zero:
\begin{equation}
\begin{aligned}
\mathbf{b}_i,\mathbf{a}_i =
\begin{cases}
0 &  i \in \mathcal{I}_p, \\
\mathbf{b}_i,\mathbf{a}_i & \text{otherwise},
\end{cases}
\end{aligned}
\end{equation}
where $i$ is the index of $i$-th rank. It means that if $i$ is in $\mathcal{I}_p$, we set both $\mathbf{a}_i$ and $\mathbf{b}_i$ of $i$-th rank to zero in preparation for subsequent expansion.

Next, more space is allocated for important ranks for better optimization. Similarly, we select the $K$ most important ranks based on $\mathbf{s}$ to form the rank index set $\mathcal{I}_e$ for expansion:
\begin{equation}
\begin{aligned}
\label{eq:Im}
    \mathcal{I}_e = \{i\ | s_i \in \text{Top}_K(\mathbf{s})\}.
\end{aligned}
\end{equation}
%\lvert I_u \rvert)\
For the pruned ranks, we assign the parameter values of the important ranks to them:
\begin{equation}
\label{copy_param}
\begin{aligned}
     \mathbf{b}_{\mathcal{I}_p},\mathbf{a}_{\mathcal{I}_p} \leftarrow \mathbf{b}_{\mathcal{I}_e}, \mathbf{a}_{\mathcal{I}_e}.
\end{aligned}
\end{equation}
Meanwhile, to ensure the stability of the optimization for the expanded ranks, the corresponding optimizer states are also assigned:
\begin{equation}
\label{copy_states}
\begin{aligned}
     \mathbf{M}_{\mathcal{I}_p},\mathbf{V}_{\mathcal{I}_p} \leftarrow \mathbf{M}_{\mathcal{I}_e}, \mathbf{V}_{\mathcal{I}_e},
\end{aligned}
\end{equation}
where $\mathbf{M}$ and $\mathbf{V}$ are the first-order and second-order moment in Adam optimizer.

However, directly assigning parameters and optimizer states from the original ranks creates a challenge: the lack of symmetry breaking between the expanded and original ranks means the optimization process is essentially trying to synchronously optimize two identical objects~\citep{DBLP:journals/corr/ChenGS15}. This makes it difficult to effectively leverage the additional capacity provided by the expanded parameter space. 
To address this issue, we propose using historical parameters and their corresponding optimizer states to break the symmetry, Eq.~\ref{copy_param} and Eq.~\ref{copy_states} change to:
\begin{equation}
     \mathbf{b}_{\mathcal{I}_p}, \mathbf{a}_{\mathcal{I}_p} \leftarrow \mathbf{b}'_{\mathcal{I}_e}, \mathbf{a}'_{\mathcal{I}_e},
\end{equation}
\begin{equation}
     \mathbf{M}_{\mathcal{I}_p}, \mathbf{V}_{\mathcal{I}_p} \leftarrow \mathbf{M}'_{\mathcal{I}_e}, \mathbf{V}'_{\mathcal{I}_e},
\end{equation}
where $\mathbf{b}'_{\mathcal{I}_e}$ and $\mathbf{a}'_{\mathcal{I}_e}$ represent the historical parameters of the important ranks, $\mathbf{M}'_{\mathcal{I}_e}$ and $\mathbf{V}'_{\mathcal{I}_e}$ represent the optimizer states\footnote{In practice, we use the parameters from half steps between the last pruning and the current pruning step.}. 
After expansion, we take the average of the corresponding expanded $s_{\mathcal{I}_p}$ and $s_{\mathcal{I}_e}$ to ensure fair competition between the expanded ranks and the original ones.

\subsection{Dynamic Top-P Threshold}
\label{top-p}
In the previous statement, we fix the number of ranks for each pruning or expansion operation to be $K$. This might overlook the actual distribution of parameter importance, potentially leading to the elimination of relatively important parameters due to quantitative constraints. 
Similar to the sampling process of text generation, we introduce the Top-P strategy~\citep{Holtzman2020The,huang-etal-2024-harder} to dynamically determine the number of operable ranks. Specifically, given the score vector $\mathbf{s}$ and a threshold $p$, we sort $s_i$ in descending order, then identify the set of operable ranks and its size as $K$:
\begin{equation}
\begin{aligned}
K = \lvert\{i \mid \sum_{j=1}^{i} s_j \geq p\}\rvert,
\end{aligned}
\end{equation}
where $i$ is the index of $i$-th rank.
A larger $p$ results in fewer ranks being operated, while a smaller one leads to more ranks being affected.

Even so, a fixed threshold $p$ still poses issues, as the learning rate decreases and the model converges, the number of ranks that need to be operated should decrease. Therefore, we design a Dynamic Top-P Threshold. To gradually reduce the number of ranks being operated, the $p$ value should progressively increase with each operation, starting from $p_{\text{init}}$ and moving towards 1 (indicating no ranks are operated). We tie this process to the learning rate scheduler used during fine-tuning to align it with the model’s learning progression. For example, given the commonly used cosine scheduler, we obtain the value of threshold $p$ at step $t$ by:
\begin{equation}
\label{get_p}
\begin{aligned}
p = p_{\text{init}} + \frac{1}{2} (1 - p_{\text{init}}) \left(1 - \cos\left(\frac{\pi t}{T}\right)\right),
% p_t = p_{\text{init}} + \frac{t}{T} \times (1 - p_{\text{init}})
% LinearLR
\end{aligned}
\end{equation}
where $T$ is the total fine-tuning steps. In implementation, we perform pruning and expansion operations every $\Delta t$ steps, which allows the LoRA module to adapt after expansion.
\begin{table*}[t]
\centering
\scalebox{0.9}{

\begin{tabular}{@{}ccccccc>{\columncolor{gray!20}}cc@{}}
\toprule
\multicolumn{1}{l}{} & \multicolumn{1}{l}{} & \multicolumn{1}{l}{} & \multicolumn{2}{c}{\textbf{Math Reasoning}} & \multicolumn{2}{c}{\textbf{Code Generation}} & \multicolumn{1}{l}{} \\ \cmidrule(lr){4-5} \cmidrule(lr){6-7}
\textbf{Model} & \textbf{Method} & \textbf{\#Params} & \textbf{GSM8K} & \textbf{MATH} & \textbf{HumanEval} & \textbf{MBPP} & \cellcolor{gray!20}\textbf{Avg.} \\ \midrule
\multirow{9}{*}{\textbf{LLaMA2-7B}} & Full-FT$^{\dagger}$ & 6738M & 66.50 & 19.80 & 38.01 & 46.03 & 42.59 \\ \cmidrule(l){2-8} 
 & LoRA & 160M & 66.31 & 19.09 & \underline{39.23} & 43.47 & \underline{42.03} \\
 & DoRA & 161M & 65.53 & \underline{19.25} & 38.41 & 42.95 & 41.54 \\
 & PiSSA & 160M & 64.87 & 17.67 & 35.77 & 39.33 & 39.41 \\
 & MiLoRA & 160M & 66.19 & 18.45 & 36.79 & 44.62 & 41.51 \\
 & ReLoRA & 160M & 62.55 & 18.08 & 35.98 & 45.59 & 40.55 \\
 & AdaLoRA & 160M & \textbf{68.04} & 17.02 & 35.16 & \textbf{46.56} & 41.70 \\
 & IncreLoRA & 160M & 65.58 & 16.93 & 34.35 & 42.77 & 39.91 \\
 & BeamLoRA & 160M & \underline{67.05} & \textbf{19.39} & \textbf{43.90} & \underline{46.30} & \textbf{44.16} \\ 
 \midrule
\multirow{6}{*}{\textbf{Mistral-7B}} & Full-FT$^{\dagger}$ & 7242M & 77.70 & 28.20 & 53.86 & 61.73 & 55.37 \\ \cmidrule(l){2-8} 
 & LoRA & 168M & 77.56 & 28.04 & \textbf{54.27} & 60.85 & 55.18 \\
 & DoRA & 169M & 77.86 & \underline{28.14} & 53.46 & 62.08 & \underline{55.39} \\
 & MiLoRA & 168M & 77.36 & 26.71 & 50.00 & \textbf{62.88} & 54.24 \\
 & AdaLoRA & 168M & \underline{77.91} & 27.53 & 46.95 & 60.14 & 53.13 \\
 & BeamLoRA & 168M & \textbf{78.11} & \textbf{28.28} & \underline{54.07} & \underline{62.70} & \textbf{55.79} \\
 \bottomrule
\end{tabular}
}
\caption{Math reasoning and code generation results for LLaMA2-7B and Mistral-7B with $r=64$ for all methods. The results are averaged over three runs. On Mistral-7B, we compare the baseline methods that perform well on LLaMA. The math reasoning results for Full-FT$^{\dagger}$ are derived from the MetaMathQA paper~\citep{yu2024metamath}.}
\label{mc}
\end{table*}
\subsection{Computational Efficiency}
Regarding fine-tuning efficiency, BeamLoRA is similar to LoRA (more details in Appendix~\ref{app-eff}), with a minimal addition of parameters in the form of a score vector $\mathbf{s}$. In terms of inference efficiency, $\mathbf{s}$ can be merged in the matrix $\mathbf{A}$: $\mathbf{A}'=\mathbf{s}\odot \mathbf{A}$, resulting in a structure identical to standard LoRA. Furthermore, the design philosophy of BeamLoRA ensures consistent ranks in various modules, allowing smooth integration with existing LoRA inference frameworks, which distinguishes it from previous works that employ varying ranks across different modules~\citep{DBLP:journals/corr/abs-2308-12043,zhang2023adaptive}.

Note that the inspiration for BeamLoRA comes from the classic Beam Search algorithm~\citep{lowerre1976harpy}, where we consider each LoRA module as a beam. 
Although BeamLoRA employs similar operations, it pursues distinct objectives. Beam Search aims to produce a single sentence to achieve the final goal, resulting in only one solution. In contrast, our approach continuously filters sub-solutions to obtain an optimal collection of sub-solutions to accomplish the objective.

\section{Experiments}
\subsection{Experimental Settings}
\noindent\textbf{Models and Datasets.}
To thoroughly evaluate our method, our experiments encompass three different base models, including LLaMA2-7B, Mistral-7B-v0.1, and LLaMA2-13B. We conduct experiments across three different domains, including math reasoning, code generation, and commonsense reasoning, utilizing a total of 12 datasets. 
For math reasoning, we fine-tune the models on the MetaMathQA dataset~\citep{yu2024metamath} and evaluate them using the GSM8K~\citep{DBLP:journals/corr/abs-2110-14168} and MATH~\citep{DBLP:journals/corr/abs-2103-03874} datasets. For code generation, we fine-tune on the CodeFeedback105K dataset~\citep{zheng2025opencodeinterpreterintegratingcodegeneration,meng2024pissa} and then evaluate using the HumanEval~\citep{DBLP:journals/corr/abs-2107-03374} and MBPP~\citep{DBLP:journals/corr/abs-2108-07732} datasets. For commonsense reasoning, we fine-tune on the Commonsense170K dataset~\citep{hu-etal-2023-llm} and evaluate on the BoolQ~\citep{clark-etal-2019-boolq}, PIQA~\citep{DBLP:journals/corr/abs-1911-11641}, SIQA~\citep{sap-etal-2019-social}, HellaSwag~\citep{zellers-etal-2019-hellaswag}, WinoGrande~\citep{DBLP:journals/corr/abs-1907-10641}, ARC-e, ARC-c~\citep{DBLP:journals/corr/abs-1803-05457}, and OBQA~\citep{mihaylov-etal-2018-suit} datasets.

\noindent\textbf{Baselines.} We compare BeamLoRA with eight baseline methods to validate the effectiveness of our proposed approach: Full-FT, LoRA~\citep{DBLP:conf/iclr/HuSWALWWC22}, DoRA~\citep{pmlr-v235-liu24bn}, ReLoRA~\citep{lialin2024relora}, PiSSA~\citep{meng2024pissa}, MiLoRA~\citep{wang2024miloraharnessingminorsingular}, AdaLoRA~\citep{zhang2023adaptive}, and IncreLoRA~\citep{DBLP:journals/corr/abs-2308-12043}. More details are presented in Appendix~\ref{sec:appendix}.

\begin{table*}[t]

\centering
\scalebox{0.8}{
\begin{tabular}{@{}cccccccccc>{\columncolor{gray!20}}cc@{}}
\toprule
\textbf{Model} & \textbf{Method} & \textbf{ARC-c} & \textbf{SIQA} & \textbf{WinoGrande} & \textbf{BoolQ} & \textbf{ARC-e} & \textbf{PIQA} & \textbf{OBQA} & \textbf{HellaSwag} & \textbf{Avg.} \\ \midrule
\multirow{8}{*}{\textbf{LLaMA2-7B}} & LoRA$^{\dagger}$ & 64.7 & \underline{79.5} & \underline{82.6} & 69.8 & 79.8 & 79.9 & 81.0 & 83.6 & 77.6 \\
 & DoRA$^{\dagger}$ & 68.2 & 76.0 & \underline{82.6} & \textbf{71.8} & \underline{83.7} & \underline{83.7} & \underline{82.4} & \underline{89.1} & \underline{79.7} \\
 & PiSSA$^{\ddagger}$ & 60.2 & 78.4 & 78.0 & 67.6 & 75.8 & 78.1 & 75.6 & 76.6 & 73.8 \\
 & MiLoRA$^{\ddagger}$ & 68.8 & \textbf{80.1} & 82.0 & 67.6 & 82.8 & \textbf{83.8} & 80.6 & 88.2 & 79.2 \\
 & ReLoRA & 59.3 & 76.9 & 77.2 & 63.9 & 75.4 & 76.4 & 63.2 & 62.2 & 69.3 \\
 & AdaLoRA & \underline{69.5} & 66.4 & 78.6 & 62.1 & \textbf{84.1} & 83.2 & 79.2 & 42.1 & 70.7 \\
 & IncreLoRA & 65.5 & 61.3 & 81.4 & 63.6 & 81.3 & 70.7 & 73.8 & 66.9 & 70.6 \\
  & BeamLoRA & \textbf{71.0} & 78.9 & \textbf{82.7} & \underline{71.6} & \underline{83.7} & 82.8 & \textbf{84.8} & \textbf{90.5} & \textbf{80.8}
 
 \\ \midrule
\multirow{5}{*}{\textbf{LLaMA2-13B}} & LoRA & \textbf{75.8} & 80.9 & 86.1 & \textbf{75.0} & 87.2 & \textbf{86.2} & 85.4 & 92.6 & \underline{83.7} \\
 & DoRA & 74.6 & \underline{81.2} & \underline{86.3} & 74.4 & 86.8 & 84.3 & 84.2 & \underline{93.4} & 83.2 \\
 & MiLoRA & 73.0 & 80.3 & \textbf{86.7} & 73.6 & 87.1 & 81.1 & \underline{85.6} & 92.0 & 82.4 \\
 & AdaLoRA & \textbf{75.8} & 73.1 & 84.5 & 67.7 & \textbf{88.3} & 83.3 & 83.4 & 90.7 & 80.9 \\
 & BeamLoRA & \underline{75.5} & \textbf{81.3} & 86.1 & \underline{74.7} & \underline{88.0} & \underline{85.6} & \textbf{86.2} & \textbf{94.3} & \textbf{84.0} \\ \bottomrule
\end{tabular}
}
\caption{Commonsense reasoning results for LLaMA2-7B and LLaMA2-13B with $r=32$ for all methods. All results with $^{\dagger}$ are taken from DoRA~\cite{pmlr-v235-liu24bn} and those with $^{\ddagger}$ are taken from MiLoRA~\cite{wang2024miloraharnessingminorsingular}.}
\label{cs}
\end{table*}
\subsection{Math Reasoning and Code Generation}
Table~\ref{mc} presents the experiments on math reasoning and code generation, demonstrating that BeamLoRA outperforms all other baseline methods in terms of overall performance. Notably, BeamLoRA not only surpasses the original LoRA across all tasks but also achieves an average performance improvement of 1.57\% compared to standard full parameter fine-tuning on LLaMA2-7B. This result is obtained while maintaining the same number of fine-tuning epochs and full data settings as the standard full parameter fine-tuning, highlighting the practicality of the BeamLoRA method. 

Furthermore, we extend our experiments on Mistral-7B by comparing BeamLoRA with the baseline methods that perform well in LLaMA2-7B experiments. The results show that BeamLoRA continues to outperform all baseline methods, surpassing Full-FT by 0.42\%. More notably, BeamLoRA shows improvement over Full-FT across all task metrics. This demonstrates that within a limited parameter budget, BeamLoRA can effectively achieve better optimization by expanding the parameter space of important ranks.

\subsection{Commonsense Reasoning}
Table~\ref{cs} presents evaluation results across eight commonsense reasoning datasets. BeamLoRA achieves the best overall performance on LLaMA2-7B, with an average accuracy improvement of 3.2\% over original LoRA and 1.1\% over the strong baseline DoRA. Similar to math and coding tasks, BeamLoRA’s performance does not rely heavily on optimal results from just two or three datasets, as observed in other baselines. Instead, it consistently performs among the top three results across most datasets, demonstrating the generalization capability of BeamLoRA. Additionally, we find that IncreLoRA and AdaLoRA are less effective in commonsense reasoning tasks, likely due to frequent rank changes across modules, which cause instability in fine-tuning. This issue is more evident in scenarios requiring extensive task evaluation.

For larger base models LLaMA2-13B, the performance gaps between different methods become smaller. In this setting, BeamLoRA still achieves a 0.3\% performance improvement over LoRA, while other methods show inferior performance compared to LoRA. This demonstrates that BeamLoRA’s approach of expanding the parameter space of important ranks is effective across base models of different sizes and can enhance the fine-tuning performance of LoRA.
% This demonstrates that BeamLoRA can effectively enhance LoRA’s performance with larger base models. 
\begin{table}[t]
\scalebox{0.93}{
\begin{tabular}{@{}lcc>{\columncolor{gray!20}}cc@{}}
\toprule
 & \textbf{GSM8K} & \textbf{MATH} & \textbf{Avg.} \\ \midrule
\textbf{ LoRA} & 66.31 & 19.09 & 42.70 \\
\textbf{ BeamLoRA} & 67.05 & 19.39 & 43.22 \\
\quad w/o Expansion & 65.88 & 18.94 & 42.41 \\
\quad w/o Assessment & 64.82 & 19.08 & 41.95 \\
\quad w/o Dynamic P. & 65.81 & 18.74 & 42.28 \\ \bottomrule
\end{tabular}
}
\caption{Results of ablation experiments. We evaluate the impact of pruning and the significance of expansion, importance assessment, and dynamic Top-P threshold in BeamLoRA.}
\label{ablation}
\end{table}
\subsection{Ablation Study}
The ablation results are shown in Table 3. Without expansion refers to only pruning the unimportant ranks, the performance experiences a slight decline compared to the original LoRA, and is markedly inferior to the complete BeamLoRA. This underscores the significance of expanding the important ranks. Without assessment refers to randomly pruning and expanding ranks, leading to a substantial performance drop. This suggests that important ranks may have been pruned, highlighting the necessity of rank assessment. Without dynamic Top-P refers to using a static operation threshold throughout fine-tuning, which also results in a performance decline. This indicates that higher thresholds should be applied during the later stages of fine-tuning, resulting in fewer ranks being operated. This approach allows the model to better adapt pruning and expansion as it converges, emphasizing the importance of dynamic thresholds.

\begin{figure}[t]
        \centering
	\subfloat[LLaMA2-7B]{
	\includegraphics[width=0.48\linewidth]{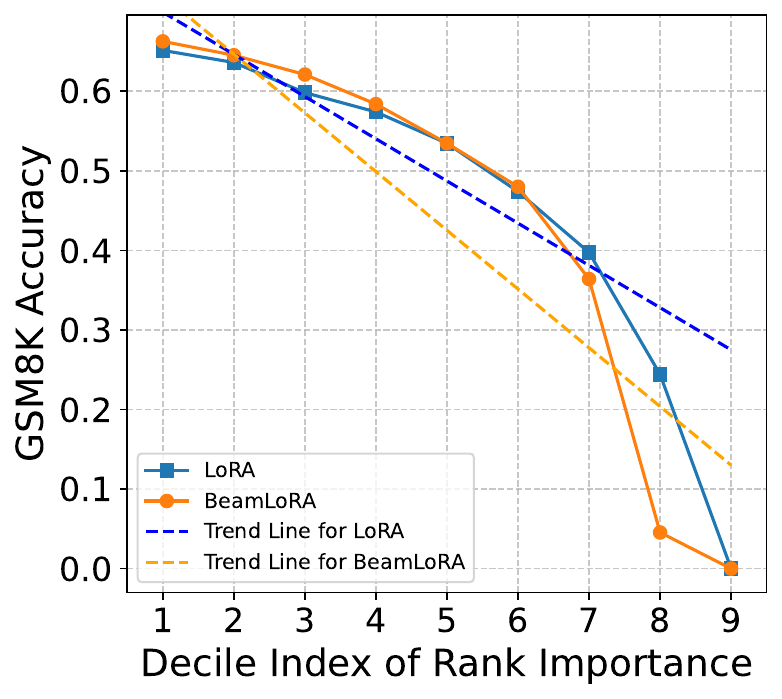}} 
        \subfloat[Mistral-7B-v0.1]{
	\includegraphics[width=0.48\linewidth]{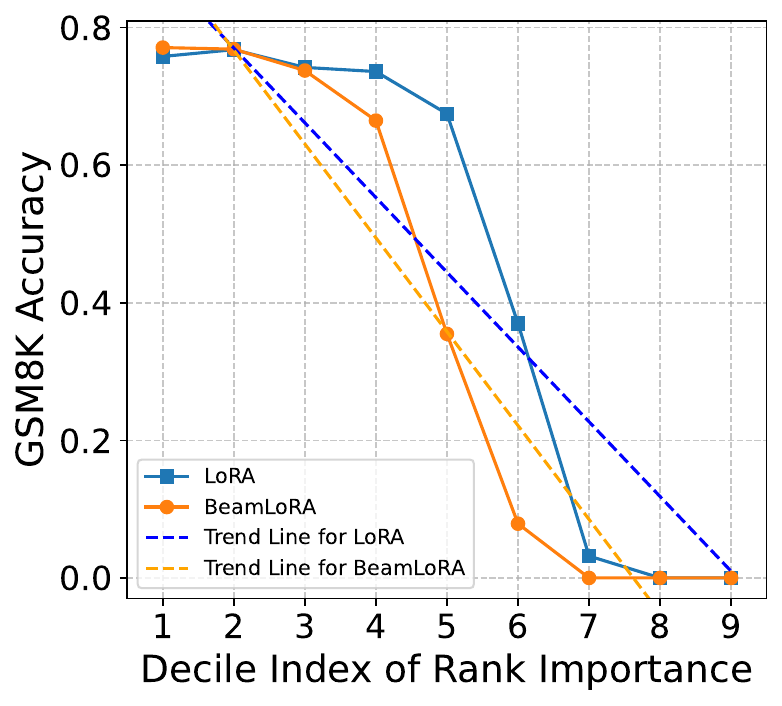}} 
\caption{Comparison of BeamLoRA and LoRA on importance of ranks within a module by pruning ranks. The solid lines represent the accuracy changes after pruning, while the dashed lines indicate the accuracy change trends caused by pruning. It can be observed that the importance is more evenly distributed among different ranks in BeamLoRA.}
\label{ana-prune}
\end{figure}
\subsection{Analysis}

\noindent\textbf{Why is BeamLoRA effective?} To further understand how BeamLoRA affects the ranks within LoRA modules, building upon our observations in Figure~\ref{PilotA}, we analyze the differences in rank importance between BeamLoRA and LoRA. As shown in Figure~\ref{ana-prune}, when pruning ranks based on their decile importance within each module, we observe that BeamLoRA’s accuracy decreases more rapidly compared to LoRA. This indicates that in BeamLoRA, the importance of different ranks within each module is more evenly distributed. Compared to LoRA, the number of important ranks increases in the BeamLoRA module, with each rank contributing more uniformly to the overall performance. This more balanced utilization of ranks explains why BeamLoRA consistently outperforms LoRA across various experimental settings.

\begin{figure}[t]
        \centering
	\subfloat[GSM8K]{
	\includegraphics[width=0.48\linewidth]{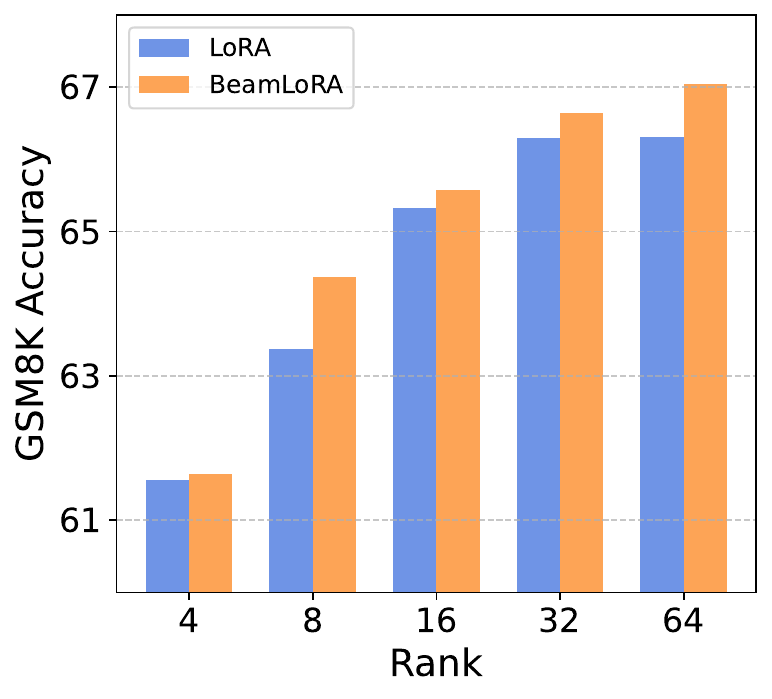}} 
        \subfloat[MATH]{
	\includegraphics[width=0.48\linewidth]{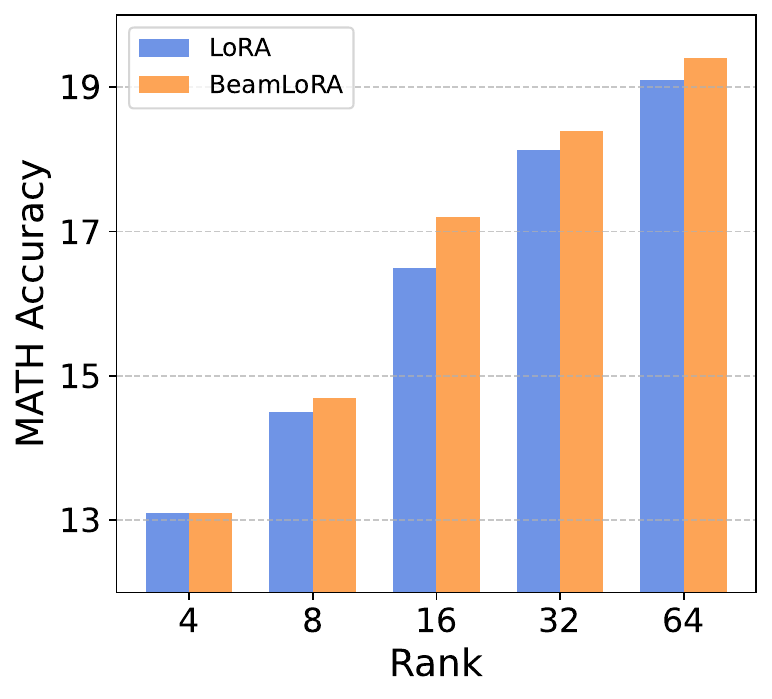}} 
\caption{Comparison of BeamLoRA and LoRA on accuracy under various rank settings on LLaMA2-7B. BeamLoRA demonstrates consistent improvements over LoRA in all settings.}
\label{ana-rank}
\end{figure}

\noindent\textbf{How does BeamLoRA perform under different rank settings?} As shown in Figure~\ref{ana-rank}. We see that BeamLoRA improves the performance of LoRA across each rank setting, demonstrating the effectiveness of BeamLoRA’s approach to compress unimportant ranks and expand important ones. In scenarios with very small ranks (e.g., $r=4$), the performance improvement brought by BeamLoRA is relatively limited compared to larger rank settings. This is because, with small rank settings and difficult tasks (e.g., Math Reasoning), LoRA is denser, leaving fewer unimportant ranks to compress and expand, thus providing a smaller operational space for BeamLoRA.

\section{Related Work}
\subsection{LoRA and Its Variants}
As one of the parameter-efficient fine-tuning methods, LoRA~\citep{DBLP:conf/iclr/HuSWALWWC22} has been widely adopted. However, it still has room for improvement in terms of accuracy.
Recent enhancements follow two main pathways: optimizing initialization and refining the fine-tuning process. For initialization, PiSSA~\citep{meng2024pissa} and MiLoRA~\citep{wang2024miloraharnessingminorsingular} apply Singular Value Decomposition on base model weights to initialize LoRA modules, with PiSSA utilizing principal singular values and MiLoRA leveraging minor ones.
For fine-tuning, DoRA~\citep{pmlr-v235-liu24bn} splits LoRA’s fine-tuning into magnitude and direction. ReLoRA~\citep{lialin2024relora} continuously merges the fine-tuned LoRA modules into the base model.  AdaLoRA~\citep{zhang2023adaptive} and IncreLoRA~\citep{DBLP:journals/corr/abs-2308-12043} optimize rank allocation across modules.
Unlike these approaches, BeamLoRA revisits the foundational aspects of LoRA and recognizes the varying importance of ranks within a module. It compresses less important ranks to free up space for expanding the important ones, allowing them to be better optimized.
\subsection{Model Pruning and Expansion}
Model pruning is typically used to remove redundant parameters in models, thereby improving efficiency~\citep{kurtic-etal-2022-optimal,ma2023llmpruner,chen-etal-2024-lemon}. Unlike previous works, our primary goal for pruning is to free up space for expanding important parameters.
Model width expansion is first introduced by Net2Net~\citep{DBLP:journals/corr/ChenGS15} and applied to CNNs. bert2BERT~\citep{chen-etal-2022-bert2bert} extends this method to the pre-training of language models, and the recent work Scaling Smart~\citep{DBLP:journals/corr/abs-2409-12903} applies width expansion to large scale base models. Unlike these approaches, we focus on parameter-efficient fine-tuning and propose compressing unimportant parameters within a limited space to expand important ones for better performance. Additionally, due to the shorter nature of the fine-tuning process than pre-training, we propose to use historical states to break symmetry in expansion, thereby ensuring fast convergence.

\section{Conclusion}
This paper introduces a PEFT method called BeamLoRA. We adopt a novel perspective to study the characteristics of ranks within a LoRA module and discover that there are differences in the importance of ranks, which change with the number of fine-tuning steps. Based on this, we propose using a dynamic threshold to prune less important ranks, freeing up space to better optimize the more important ones. Extensive experiments demonstrate that BeamLoRA effectively enhances the performance of LoRA across different base models, tasks, and settings, outperforming other baseline methods. 
\section*{Limitations}
BeamLoRA introduces a method that compresses less important ranks while expanding important ones during fine-tuning. Although this approach achieves good performance in parameter-efficient fine-tuning, the existing implementation requires the addition of a trainable assessment vector $\mathbf{s}$ over ranks. In the context of full-parameter training, a full-rank matrix does not have the low-rank structure like the $\mathbf{B}$-$\mathbf{A}$ decomposition in LoRA, making it impossible to add the vector $\mathbf{s}$. How to extend this method to full-parameter training scenarios remains an area for future research exploration.
\section*{Acknowledgments}
The authors thank Yilong Chen from the Institute of Information Engineering for the help. This work is supported by the National Natural Science Foundation of China (No. 62472419, 62472420).

% Bibliography entries for the entire Anthology, followed by custom entries
\bibliography{anthology,custom}
% Custom bibliography entries only
%\bibliography{custom}
\newpage
\appendix
\section{Experimental Setup}
\label{sec:appendix}
\subsection{Baselines}
We select several baselines to verify the effectiveness of our method. \textbf{Full-FT} fine-tunes all model parameters, delivers strong performance but requires substantial computational resources. \textbf{LoRA}~\citep{DBLP:conf/iclr/HuSWALWWC22} is one of the most efficient PEFT methods, offering computational efficiency, though its accuracy often differs from full parameter fine-tuning. \textbf{DoRA}~\citep{pmlr-v235-liu24bn} decomposes LoRA's fine-tuning into magnitude and direction components. \textbf{ReLoRA}~\citep{lialin2024relora} continuously merges the obtained LoRA parameters into the base model during fine-tuning. \textbf{PiSSA}~\citep{meng2024pissa} and \textbf{MiLoRA}~\citep{wang2024miloraharnessingminorsingular} perform Singular Value Decomposition (SVD) on the base model; PiSSA fine-tunes the significant components of the decomposition, while MiLoRA fine-tunes the minor components. \textbf{AdaLoRA}~\citep{zhang2023adaptive} begins fine-tuning with a rank setting higher than the target and prunes redundant ranks during the process to achieve optimal rank allocation across different modules. In contrast, \textbf{IncreLoRA}~\citep{DBLP:journals/corr/abs-2308-12043} starts with a rank setting lower than the target and progressively increases the rank during fine-tuning to achieve optimal rank allocation across modules.
\subsection{Implementation Details}
In the math and code tasks, we follow \citet{yu2024metamath} and set the Full-FT learning rate for LLaMA2-7B to 2e-5 and for Mistral-7B to 5e-6, with a batch size of 128. The models are fine-tuned for 3 epochs on the dataset. For the ReLoRA method, we use the same settings as Full-FT. For other PEFT methods, we add the LoRA module to all linear layers. Following \citet{wang2024miloraharnessingminorsingular} we set their learning rate to 3e-4 on LLaMA2-7B and 6e-5 on Mistral-7B, with a batch size of 32 and fine-tuning for 3 epochs. Our evaluations are conducted using the MetaMathQA and the evalplus~\cite{evalplus} codebase.

In the commonsense reasoning tasks, to facilitate comparison, we follow the approach of \citet{hu2023llm} and \citet{pmlr-v235-liu24bn} by adding LoRA to q\_proj, k\_proj, v\_proj, up\_proj, and down\_proj modules. We use the optimal learning rate from \{2e-4,3e-4\} for different methods on LLaMA2-7B, with a batch size of 16 and fine-tuning for 3 epochs. For the larger LLaMA2-13B, following \citet{pmlr-v235-liu24bn}, we reduce the learning rate to the optimal \{1e-4,2e-4,3e-4\} for different methods, while keeping other settings unchanged. These settings are applied to all PEFT methods. Our evaluations are conducted using the LLM-Adapters~\cite{hu-etal-2023-llm} codebase. All our experiments are conducted on four H800 GPUs.

The implementation of BeamLoRA mainly includes assessment, pruning-expansion, and dynamic Top-P threshold. The most crucial pruning-expansion is uniformly set in the first two epochs to facilitate. BeamLoRA introduces two hyperparameters: the initial value of the dynamic threshold $p_{\text{init}}$ and the operation interval step $\Delta t$. We analyze their impact in Section~\ref{hyperparameter-ana}. The detailed settings can be found in Table~\ref{hyper-table}.
\begin{figure}[t]
        \centering
	\subfloat[Number of Operations]{
	\includegraphics[width=0.48\linewidth]{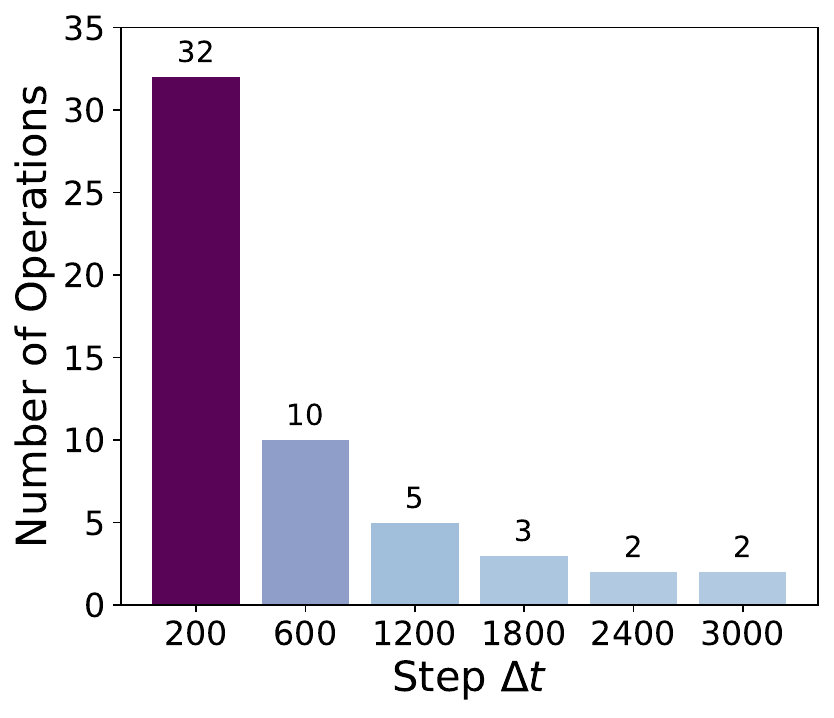}} 
        \subfloat[Code Performance]{
	\includegraphics[width=0.48\linewidth]{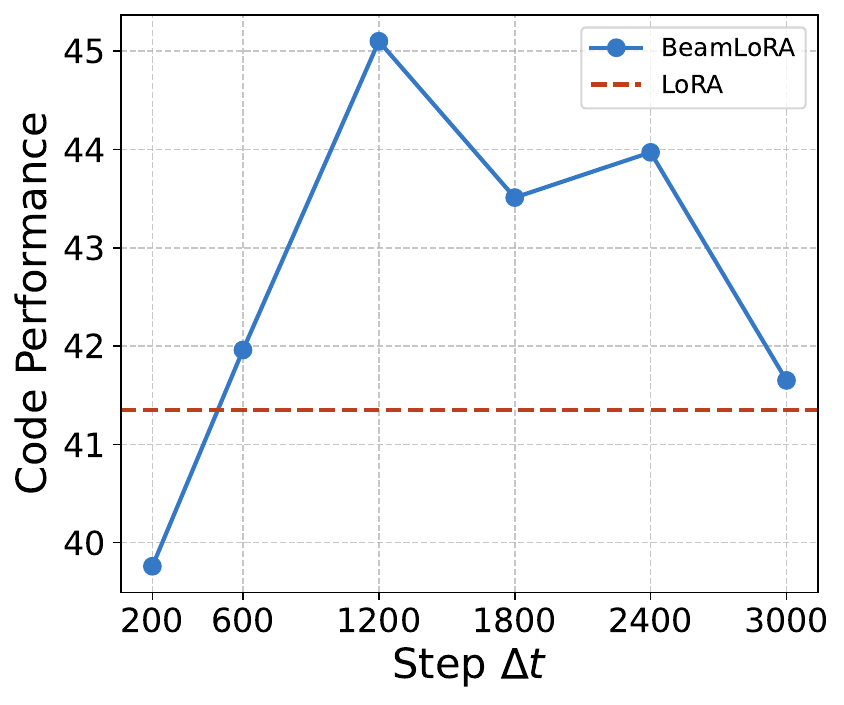}} 
\caption{Effect of step $\Delta t$. We set $p_{\text{init}}$ to 0.95 and take the average performance across the code tasks on LLaMA2-7B with $r=64$.}
\label{app-step}
\end{figure}
\begin{figure}[t]
        \centering
	\subfloat[Number of Operated Ranks]{
	\includegraphics[width=0.48\linewidth]{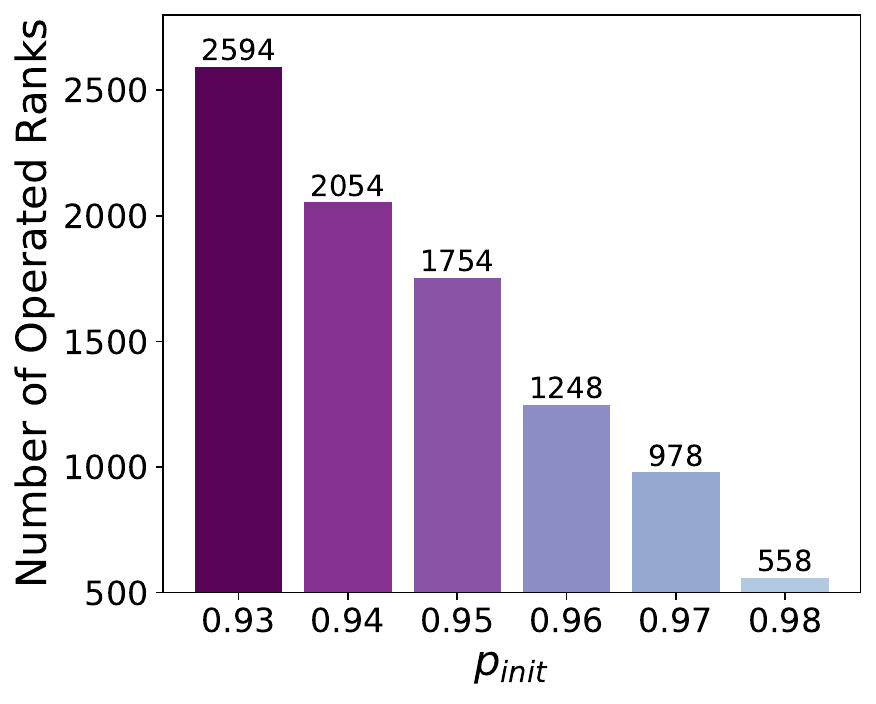}} 
        \subfloat[Code Performance]{
	\includegraphics[width=0.48\linewidth]{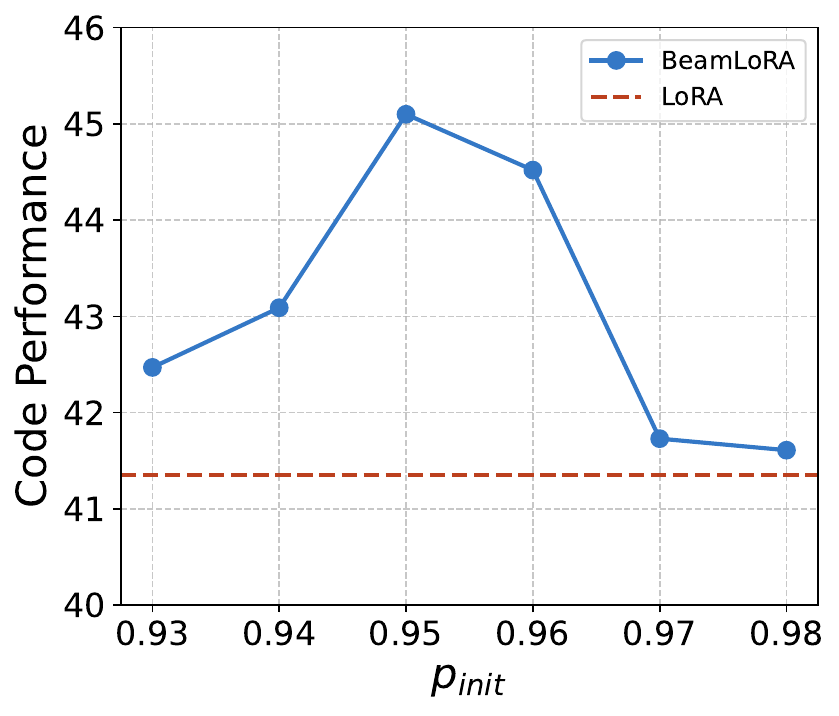}} 
\caption{Effect of $p_{\text{init}}$. We set the step $\Delta t$ to 1200 and take the average performance across the code tasks on LLaMA2-7B with $r=64$.}
\label{app-p}
\end{figure}
\section{Additional Experiment Results}
\subsection{Analysis of Hyperparameters}
\label{hyperparameter-ana}
\begin{table*}[ht]
\centering
\scalebox{0.9}{

\begin{tabular}{@{}ccccccc@{}}
\toprule
 & \multicolumn{2}{c}{Math Reasoning} & \multicolumn{2}{c}{Code Generation} & \multicolumn{2}{c}{Commonsense Reasoning} \\ \midrule
Base Model & LLaMA2-7B & Mistral-7B & LLaMA2-7B & Mistral-7B & LLaMA2-7B & LLaMA2-13B \\
Rank r & 64 & 64 & 64 & 64 & 32 & 32 \\
$p_{\text{init}}$ & 0.95 & 0.95 & 0.95 & 0.95 & 0.96 & 0.96 \\
Step $\Delta t$ & 3000 & 2500 & 1200 & 800 & 3200 & 2400 \\ 

\bottomrule
\end{tabular}
}
\caption{Detailed settings of hyperparameters in BeamLoRA. In order to thoroughly evaluate the method’s performance, we utilize two standard configurations in our experiments: $r=64$ and $r=32$. The $p_\text{init}$ is based on the rank size. $\Delta t$ is determined based on the total number of fine-tuning steps across different datasets to control the number of operations.}
\label{hyper-table}
\end{table*}

\noindent\textbf{Effect of step $\Delta t$.}
The hyperparameter $\Delta t$ determines the number of adaptation steps the model takes after each pruning-expansion operation. As shown in Figure~\ref{app-step}, when $\Delta t$ is too small (e.g., 200), the model cannot quickly adapt to the pruning and expansion due to the excessive number of operations, resulting in suboptimal performance. As $\Delta t$ increases, performance gradually improves, reaching its peak at $n=1200$ with five pruning-expansion operations. When $\Delta t$ continues to increase, the number of operations becomes too few, diminishing the benefits of the pruning-expansion operation. In our implementation, we determine the number of BeamLoRA operations based on the total number of fine-tuning steps. Specifically, for the MetaMathQA dataset, we select from \{2500, 3000, 3500\}; for the CodeFeedback dataset, we select from \{800, 1000, 1200\}; and for the Commonsense dataset, we select from \{2400, 2800, 3200\} for $\Delta t$.

\noindent\textbf{Effect of $p_{\text{init}}.$} The hyperparameter $p_{\text{init}}$ determines the number of affected ranks in each pruning-expansion operation. According to Eq.~\ref{get_p}, a smaller $p_{\text{init}}$ results in more ranks being operated. As shown in Figure~\ref{app-p}, we see that when a larger number of ranks is operated (e.g., $p=0.93$), the performance, while still better than LoRA, is not optimal. As $p_{\text{init}}$ increases, the number of operated ranks decreases, reaching optimal performance at $p=0.95$. However, if $p_{\text{init}}$ continues to increase, the number of operated ranks further decreases, reducing the advantages of the pruning-expansion process, and performance gradually declines to levels comparable to LoRA.

\begin{figure}[t]
        \centering
	\includegraphics[width=1\linewidth]{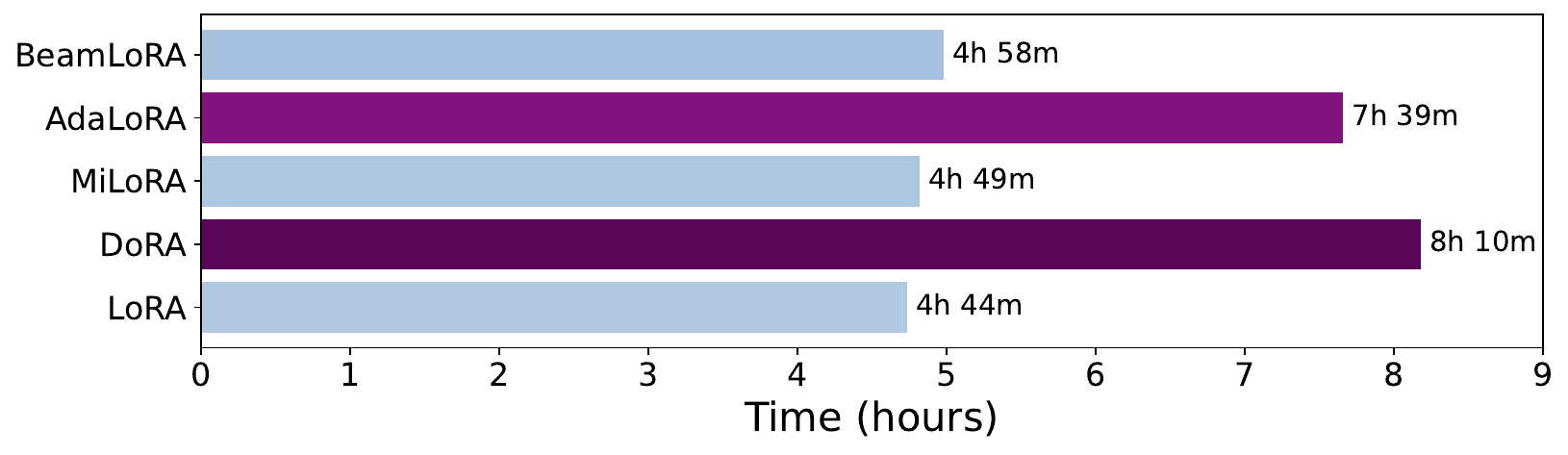}
    \caption{Fine-tuning time for different methods on LLaMA2-7B with $r=64$.}
\label{app-time}
\end{figure}
\begin{figure}[t]
        \centering
	\includegraphics[width=0.86\linewidth]{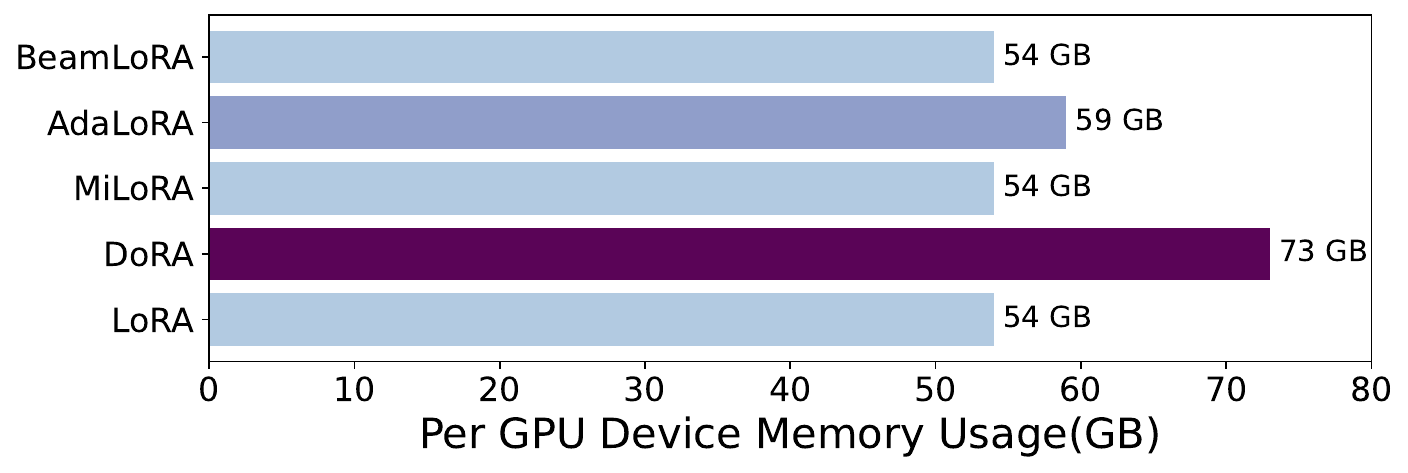}
    \caption{Fine-tuning memory usage for different methods on LLaMA2-7B with $r=64$.}
\label{app-memory}
\end{figure}

\subsection{Fine-tuning Efficiency}
\label{app-eff}
The efficiency of fine-tuning primarily involves two aspects: fine-tuning time and memory usage. We present the fine-tuning time for different methods on the MetaMathQA dataset in Figure~\ref{app-time}. We see that BeamLoRA and MiLoRA require a similar time as LoRA. However, DoRA and AdaLoRA require 1.73 times and 1.62 times the fine-tuning time of LoRA, respectively, thereby losing some of the time-saving advantages that LoRA offers.

In terms of memory usage, as shown in Figure~\ref{app-memory},  it can be observed that BeamLoRA and MiLoRA also have similar memory usage on each GPU as LoRA. AdaLoRA requires slightly more memory compared to these three, while DoRA requires 1.35 times the memory of LoRA. Overall, in terms of fine-tuning time and memory usage, BeamLoRA is similar to LoRA and significantly lower than DoRA and AdaLoRA, rendering it practical.

\end{document}